# Real-time Chatbot Romance Scam Detection in Stage Classification

Yi-cheng Liu [1], Guan-ru Lai [2], Yi-cheng Lai [3], Lyn Chao-ling Chen (Corresponding) [4, *]
Department of Computer Science, National Chengchi University, Taipei, Taiwan [1],
Graduate Institute of Artificial Intelligence Cross-disciplinary Technology, National Taiwan University of Science and Technology, Taipei, Taiwan [2],
Institute of Information Science, Academia Sinica, Taipei, Taiwan [3],
Interdisciplinary Artificial Intelligence Center, National Chengchi University, Taipei, Taiwan [4],
louisliu911231@gmail.com [1], M11552023@mail.ntust.edu.tw [2], lai0017@as.edu.tw [3], lynchen@ntu.edu.tw [4, *]

## Abstract

**Romance scam has become a serious problem via social media, and the improvement of artificial intelligence enable to fake identify easily. In the study, a chatbot has developed to detect romance scam in real-time through Line Messaging API. Using the seven stage theories in the social science field, the stages of romance scam can be detected through large language model. In addition, once a romance scam is detected, the chatbot provides explanations of the stage and prevention advices through large language model. The contribution of the work provides a practical application, and can be used in real scenario for reducing the risk of romance scam.**

## I. Introduction

In the study, the issue of romance scan has been discussed. Romance scan often happened on social media, the scammers fake online identities to gain the affection and trust of the victims, and use the illusion of love or intimacy to manipulate them or cheat their money [1]. The announcement of dating apps such as Tinder, Bumble and Facebook Dating changes the way people pursue relationships, and they are more easily to become victims of online romance scams [2] [3]. In Taiwan, more than 50% of people encounter fraud every week, almost 30% of people pay money or provide personal information within one hour in 2024 [4]. Scams on social media increase most with 21% compared to 2023, and becomes the second fraud type than the telephone fraud in Taiwan [4]. Scammers often create an image of love, romance and commitment on social media to become more attractive, and also create multiple accounts on different dating platforms to increase the chances of success [2] [5]. Some scammers even donate to charities for concealing their crimes [6]. For the development of artificial intelligence, scammers commit crimes quickly to generate fraudulent content through FraudGPT or WormGPT, and the elders and woman have high risk of romance scan than the others [7] [8]. Elders tend to share personal information online, and become the major group of victims than the other ages [9] [10]. In addition, many users show their privacy and personal information publicly on social media or online forums, and have great chance to encounter identity theft and other cybercrime scams [11]. Victims often disagree to disclose romance scam for embarrassment, or not seek support from the others for the fear of disapproval, and the depression grief, or Post-Traumatic Stress Disorder (PTSD) harm their mental health [12]. Hence, a chatbot has developed for detection of romance scan in real-time in the work.

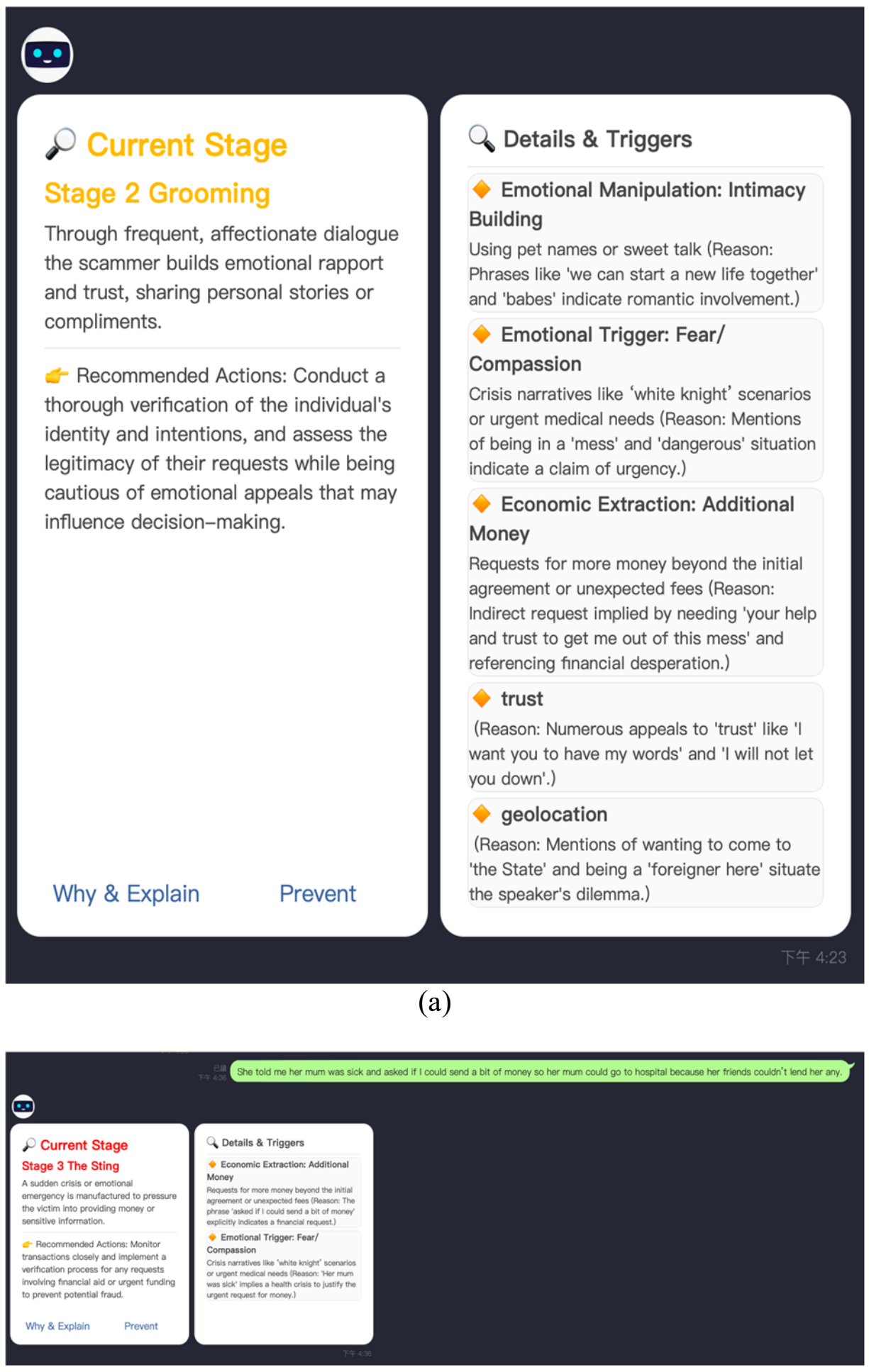

(a)

(b)

Figure 1. Real-time romance scam chatbot showing the stage number, name and a brief description corresponding to the scam stage: (a) long text input, (b) single sentence input.

This paper is structured as follows: Section 2 reviews the theories of romance scam in social science, and technologies for romance scam prevention. Section 3 outlines the methodology that introduces the system architecture and the interface design. Section 4 evaluates the chatbot and discusses the evaluation results. Section 5 concludes the study.

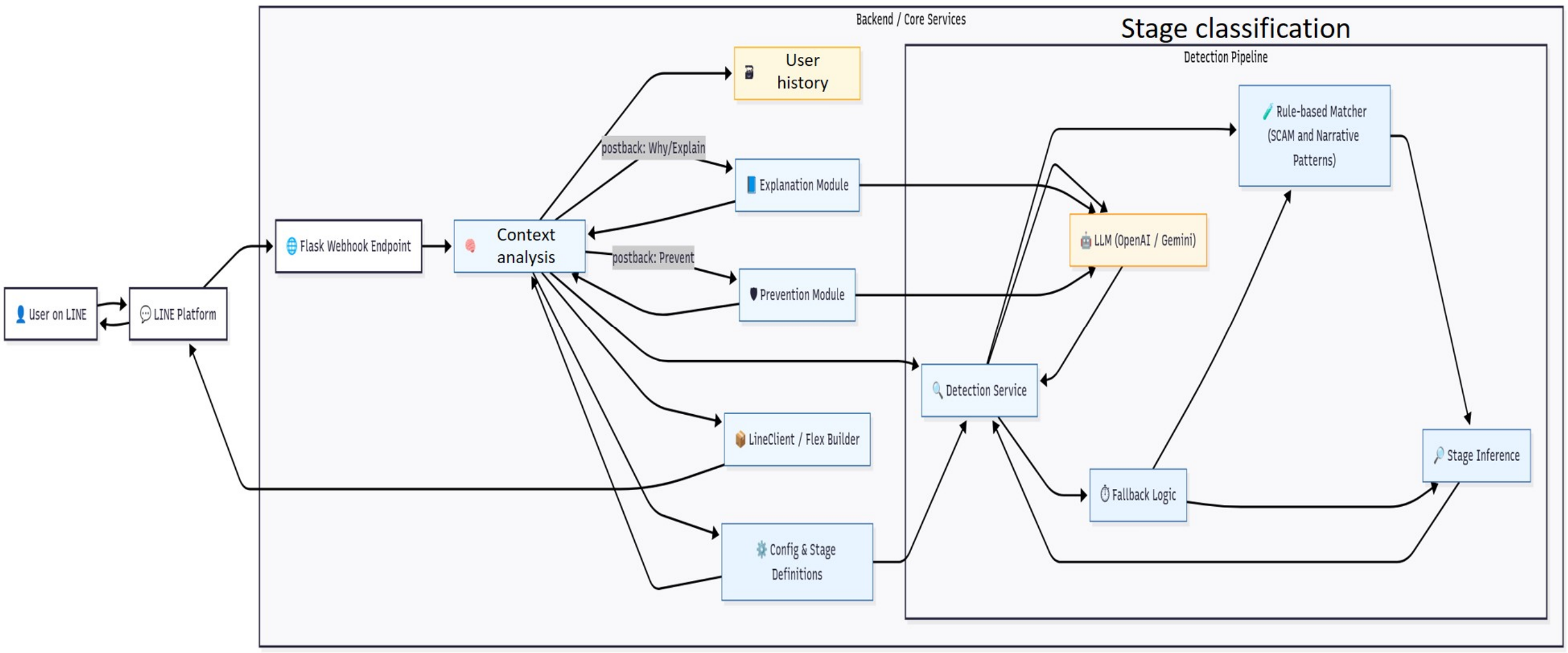

Figure 2. System architecture of romance scam chatbot.

## II. RELATED WORKS

Stages of online romance scan has been defined in the researches of romance scan field. Four stages of Andoh-Baidoo include client recruitment stage that scammers recruit people who seek get-rich-quick on internet cafes or online platforms, client grooming stage that scammers trains the new employees to create multiple attractive fake accounts and profiles for creating romance relationship and gaining trust of victims trough online chatting, initial extortion stage that scammers make small amount of financial demands from victims or threats them to explore the romance scan, and escalated extortion stage that scammers make large amount of financial demands from victims [6]. Seven stages of Whitty include the stage of motivated to find the ideal partner that scammers seek the victims who peruse love online, the stage of presented with the ideal profile that scammers fake identities to win quick favor, the grooming process stage that scammers create deep relationship with intimate and emotional online chatting frequently, the sting stage that scammers fake emergent crises of money requests when they already gained the trust of the victims, the stage of continuation of the scam that scammers fakes a series of new emergencies for repeating payments of victims, the sexual abuse stage that scammers threat victims for further funds or favors when they obtain the private images or messages of victims, the re-victimisation (re-victimization) stage that scammers of new scammers pretend as investigators or refund agents to get payment from victims continually [13].

For prevention of romance scan, an architecture has designed for sending messages to warm users on a dating platform using contextual analysis [14]. Natural Language Processing (NLP) helps to filter user profiles of scammers from general users using meaning and context information [2]. Combination of image, profile description and structured data analysis can identify suspicious online accounts precisely in the registration stage [15]. Hence, in the study, Whitty stage model was adopted for more stages and details of definitions than the other models, and evaluates the stages of romance scam through Large Language Model (LLM).

## III. SYSTEM DESIGN

### A. Line Chatbot Real-time Detection

System architecture of the chatbot romance scan detection in real-time includes context analysis and stage classification (implemented in Python Flask and Line Messaging API) (Figure 2). From the Line messages of user input, given an input of texts, the webhook receiver on local device receives the Line webhook and the texts, and extracts the user ID, texts and token for message signature authentication. The system notices the users of recording and analysis of texts messages in advance for privacy concern. Using the record of text messages in user history, the Generative Pre-trained Transformer 4 Omni model (GPT-4o) model was adopted for context analysis, and the advantage of the LMM model is that the semantic features of romance scam can be identified efficiently. According to the results of context analysis, the webhook receiver call the message process to determine the stage of the romance scam in stage classification. GPT-4o model was also adopted for providing explanations and prevention advice to the users (Figure 3 (a) and (b)), and the webhook receiver determines whether call the explanation mode or prevention mode when receives a postback from user click of “Why & Explain” or “Prevention Tips” (Figure 1 (a)). Chatbot responses to the users using a predefined template and a text style in the form of a stage number, a stage name, and a list of triggered factors according to the seven stages with brief descriptions in the Line chatting.

In addition, the generated texts in explanation mode are in paragraphs, and the listing prevention advice differs in different stages both considering the readability. Otherwise, when no romance scam is detected, a flag of "no risk" is returned.

## B. *Large Language Model in Stage Classification*

Using the Whitty stage model in the stage definition of romance scam [13], the record of 7 stages includes the stage ID (0 to 6) as follows: stage 0: motivated to find the ideal partner, stage 1: presented with the ideal profile, stage 2: presented with the ideal profile, stage 3: the grooming process, stage 4: the sting, stage 5: continuation of the scam, stage 6: re-victimisation (re-victimization).

GPT-4o model was adopted for analyzing the scam stage of the Line messages, and the prompt input for context analysis of user text history is as follows: *"You are an expert romance-scam detection assistant based on Witty et al.'s seven-stage framework. Given a user message (which may be either a pure dialogue snippet or a user describing their personal experience), do the following: 1. Decide whether the message is a personal experience description or pure dialogue. ("experience" vs "dialogue") 2. Identify which scam stage the message best fits into. Use the integer mapping: 0: Find the Dream Mate 1: Contact via fake profile 2: Grooming 3: The Sting 4: Continuation of the Scam 5: Sexual Exploitation 6: Re-victimization 3. Extract all relevant scam-related labels/triggers present (e.g., payment, urgency, identity_inconsistency, unverified_communication, romance, authority, crisis, money_request, sexual_request, friendly, geolocation, etc.). 4. For each of the above (input_type, each label, and stage), provide a concise rationale citing exact phrases, contradictions, or patterns in the message that justify your classification."*

The output of the LLM results is in a customized format. For example, from an input of text messages of "She told me her mum was sick and asked if I could send a bit of money so her mum could go to hospital because her friends couldn't lend her any," the output results is "input_type: dialogue, narrative_reason: No narrative patterns matched, stage: 3, labels: [money_request, crisis, authority], rationale: {input_type: Detected personal experience: the user describes being personally asked for money ('She told me...' and 'asked if I could send')., labels: {money_request: The request to 'send a bit of money' clearly indicates a financial demand., crisis: The reason provided for the money request, 'her mum was sick... so her mum could go to hospital,' frames a situation of urgency and crisis., authority: Reference to 'her friends couldn't lend her' implies social pressure and lack of available support, leveraging social trust.}, stage: A crisis and money request from a fake relationship fits into 'The Sting' (stage 3).}." The list of keywords showing why the text input is classified in the scam stage (Figure 1 (a) and (b)). For example, keyword of "emotional manipulation" uses words such as "we can start a new life together" or "babes" implying a romantic relationship with the victim, keyword of "emotional trigger" contains urgent pleas or tragic stories such as "in trouble" or "in danger" to evoke sympathy of the victim, keyword of "economic extraction" uses words such as "I need your help" or implied financial demands to the victim, keyword of "Trust" uses the word of trust repeatedly for emphasizing the impressions of the victim, and

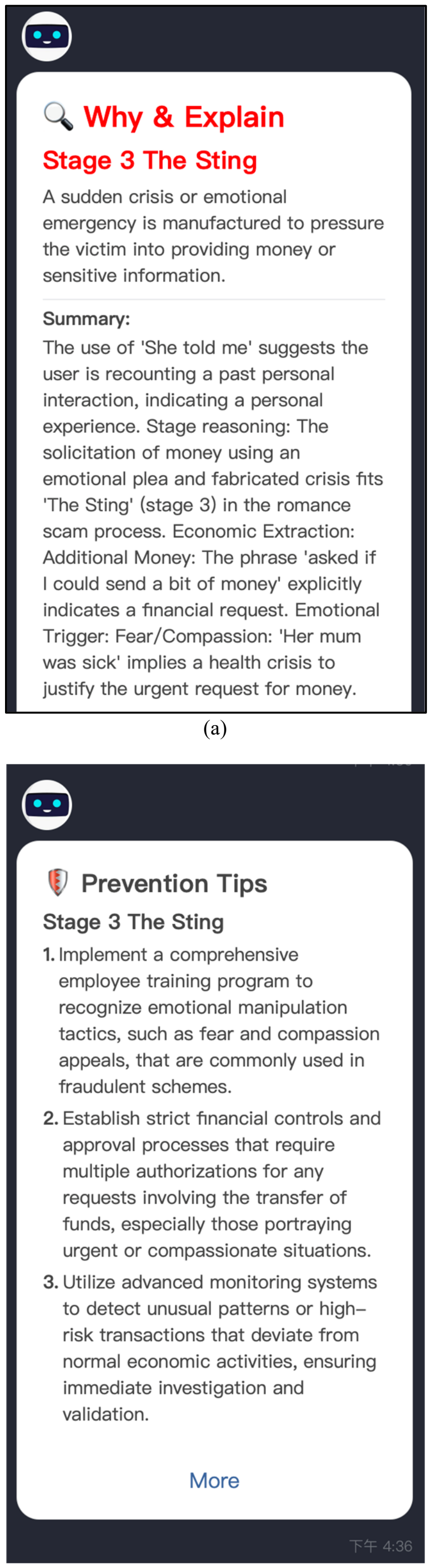

(a)

(b)

Figure 3. Single sentence input: (a) explanation module: explanations of the scam stage, and (b) prevention module: prevention advice of the scam stage.

TABLE I. EVALUATION OF INDIVIDUAL ROMANCE SCAM STAGE WITHOUT CONTEXT INFORMATION

| Stage | Name | Correct Judgement | Incorrect Judgment | Total Samples | Accuracy (%) |
|---|---|---|---|---|---|
| 0 | Motivated to find the ideal partner | 9 | 3 | 12 | 75 |
| 1 | Presented with the ideal profile | 9 | 4 | 13 | 69.2 |
| 2 | The grooming process | 43 | 15 | 58 | 74.1 |
| 3 | The sting | 8 | 20 | 28 | 28.6 |
| 4 | Continuation of the scam | 13 | 1 | 14 | 92.9 |
| 5 | The sexual abuse | 4 | 0 | 4 | 100 |
| 6 | The re-victimisation (re-victimization) | 14 | 9 | 23 | 60.8 |

keyword of "geolocation" mentions a foregoing country or the need of visiting the country of the victim. Hence, in the study, using context information, large language model was adopted to detect romance scam effectively.

### C. Large Language Model in Explanation Module and Prevention Advice Module

For generating explanations of the scam stage, the GPT-4o model was adopted, and the prompt input for generating the reasons of the input texts and the scam stage are as follows: *"I just detected a message, classified as stage {stage_num} ({stage_name_for_explain}), the trigger factors are {trigger_factors}. Please use 2 to 3 sentences to briefly explain why you made such a judgment,"* and the prompt input for proving more information for users to understand of the scam stage is as follows: *"I previously classified a message as stage {stage_num} ({stage_name}) with input type '{input_type}'. The detected triggers are: {trigger_names}. Existing brief rationale: input_type: {rationale.get('input_type','')}; stage: {rationale.get('stage','')}; labels: {rationale.get('labels',{})}. Please provide a more detailed explanation (4-5 sentences) of why this classification makes sense, including how the combination of signals supports the stage and any caveats or uncertainties."* The variables consist of the prompt include the necessary information of scam stage (stage_num), stage name (stage_name_for_explain), predefined triggered factors of seven stages (trigger_factors), transformation of triggered factors (trigger_names), long text input or single sentence input (input_type) and return interpretations from LLM (rationale.get('input_type','')) (rationale.get('stage','')) (rationale.get('labels',{})).

For providing prevention advice to reduce the risk of romance scam, the GPT-4o model was adopted, and the prompt input for generating prevention advice of the scam stage is as follows: "Given fraud stage {stage_num} ({stage_name}) and trigger factors {trigger_names}, provide exactly three practical prevention suggestions. Return them as a numbered list, each in one sentence," and the prompt input for proving more prevention information of the scam stage for users is as follows: *"Earlier you gave three concise prevention suggestions for fraud stage {stage_num} ({stage_name}) based on the trigger factors: {trigger_names}:\n {summary_list_text}\n Please expand each of the three suggestions into a detailed explanation. For each one, explain why it's important and how to implement it in practice, using 2 to 3 sentences per item. Keep the numbering (1., 2., 3.) and return only a numbered list; do not add extra introductory or closing paragraphs."* The variables consist of the prompt include the necessary information of scam stage (stage_num), stage name (stage_name_for_prevent), predefined triggered factors of seven stages (trigger_factors), transformation of triggered factors (trigger_names), long text input or single sentence input (input_type) and list of the 3 prevention advice for generating further information (summary_list_text). The response time is tolerable within few seconds without delays. Hence, the explanation module helps users to understand the reasons of the detected scam stage, and the prevention module helps to reduce the risk of romance scam.

### D. Visual Feedback in Interface Design

In Line chatting, user input includes two data types according to the length of the texts: long text input (Figure 1 (a)) and single sentence input (Figure 1 (b)) that enhances of the capability of the chatbot. The long text input represents the entire conversations or multiple content, and the single sentence input (Figure 1 (b)) means short messages. By considering the readability, the design of visual feedback uses different colors in the seven stages, showing the corresponding colors in the stage name and the icons with keywords of the detected scam stages (Figure 1 (a) and (b)). For example, in stage 2 uses orange-yellow color (Figure 1 (a)), and in stage 3 (The sting) *uses red color* (Figure 1 (b)). Hence, the visual feedback helps users to distinguish different scam stages.

## IV. EVALUATION AND DISCUSSIONS

Real romance scam data (Data number: 152) were collected from the Reddit website and were labeled with seven stages of the romance scam manually for evaluation [16] [17] [18] [19] [20] [21] [22] [23] [24] [25] [26] [27] [28] [29] [30] [31]. From the detection results of the chatbot (Table 1), in identifying the stages of the romance scam have high accuracies in the stages of "The sexual abuse" (100%) and "Continuation of the scam" (92.9%). For "The sexual abuse" stage involves clear features such as sexual inducements or request of nude photos that the chatbot classified correctly, and the "Continuation of the scam" also have clear features such as urgent assistance or money request, for example "I was in the most vulnerable position when this guy swooped and exploited and violated me" and "We had a lot of sexting while I fed and supported her while she was out of town with a new temp job" in correct classification of "The sexual abuse" stage, and "Maria claimed she was progressing paperwork at the embassy and needed the remaining $2300 to complete it, or they'd have to restart the process" and "Last year, someone hacked into her Chase bank account and received every payment of her bills for four months" in correct classification of the "Continuation of the scam" stage. Wrong classification happens when the data have obvious features of other stages in the condition without context information, for example the data of "Will has sent Maria over $30,000 over the last seven years" in the "Continuation of the scam" stage was identified as the "The re-victimisation (re-victimization)" stage.

In the stages of the "Motivated to find the ideal partner" (75%), "The grooming process" (74.1%), "Presented with the ideal profile" (69.2%) and "The re-victimisation (re-victimization)" (60.8%) also have moderate accuracies (Table 1). The data in the correct classification with semantic meanings match the stags, for example "My grandmother met a man on Facebook Dating" and "He apparently met a 'former actress' online after my parents' divorce" in the "Motivated to find the ideal partner" stage, "Will fell in love with Maria after they began video chatting and calling each other" and "I think it's going to be difficult to convince her parents to let her travel to a foreign country if we're not married so maybe we will do the fake marriage thing but we can worry about that later" in "The grooming process" stage, "He sent a screenshot of my location when I was going to a beach with a friend to make it seem like he knew me" and "She was contacted on FB by a close friend's younger brother who she has known for over 30 years" in the "Presented with the ideal profile" stage, and "But somehow I got put in that damn position again" and "Despite being shown red flags for years, Will continues to send her money, unable to see that he's being exploited" in the "Presented with the ideal profile" stage. The similar situations caused by the lack of context information, for example the data of "My friend has fallen for a love scammer two years ago" in the "Motivated to find the ideal partner" stage was identified as the "Continuation of the scam" stage, the data of "We had been talking on Hangout; I asked her if we could talk on WebCam—she said it was getting fixed; I asked for a call—she said they were not allowed to have phones" in "The grooming process" stage was identified as "The sting" stage, the data of "She met a man there who was very handsome and they have been speaking ever since" in the "Presented with the ideal profile" stage was identified as "The grooming process" stage, the data of "When she asks for her money back, he says he'll 'Zelle her' but doesn't, then claims he gave the money to his kids" in "The re-victimisation (re-victimization)" stage was identified as the "Continuation of the scam" stage.

In "The sting" stage, poor performance (28.6%) occurred for only input of single stage data in the evaluation (Table 1), and the wrong classification data were all (Data number: 20) misclassified to the "Continuation of the scam" stage for lack of context information. Correct classification in "The sting" stage mentions keywords of money or in danger directly such as "She told me her mum was sick and asked if I could send a bit of money so her mum could go to hospital because her friends couldn't lend her any" and "He was in a jam and started reaching out to people to generate funds to pay 'taxes' as his fake soulmate was threatening to kill herself." The wrong classification in "The sting" stage occurs for the data without direct information in the condition without context information, for example the data of "She wanted me to send more money as they wouldn't allow access to money or their bank details" and "About 4 weeks ago her phone broke in the shower and she asked me to replace it" were identified as the "Continuation of the scam" stage. The evaluation results show that the stage has more clear features with better performance in the condition without context information. Hence, in the study, the chatbot detects romance scam relies on context information in real scenario, and definitely performs better than the evaluation results.

## V. Conclusion

The chatbot provides a useful application for reducing the risk of romance scam, and can be applied in real scenario to provide real-time detection in Line chatting. The explanations and the prevention advice of the detected scam stage also provide useful information to understand the stages of romance scam and to alert with any suspicious chatting on social media.

## Acknowledgments

This work was partially supported by the National Science and Technology Council, Taiwan, under grants 114-2635-E-004 -001-. Special thanks to the reporter, Bo-wei Hsu.